\documentclass{article}

\usepackage[T1]{fontenc}         
\usepackage{lmodern}             
\DeclareFontFamily{T1}{pcr}{}
\DeclareFontShape{T1}{pcr}{m}{n}{<->ssub*lmtt/m/n}{}
\DeclareFontShape{T1}{pcr}{m}{it}{<->ssub*lmtt/m/it}{}
\DeclareFontShape{T1}{pcr}{b}{n}{<->ssub*lmtt/b/n}{}
\usepackage{microtype}
\usepackage{graphicx}
\usepackage{booktabs}
\usepackage{hyperref}

\usepackage[accepted]{icml2025}

\usepackage{amsmath,amssymb,amsthm}
\usepackage{mathtools}

\usepackage{multirow}
\usepackage{array}
\usepackage{subcaption}
\usepackage{xcolor}
\usepackage{colortbl}
\usepackage{caption}
\definecolor{tableheader}{RGB}{255,178,102}  
\usepackage{tikz}
\usetikzlibrary{arrows.meta,positioning}

\newcommand{\la}{\lambda_a}
\newcommand{\lc}{\lambda_c}
\newcommand{\Lalign}{\mathcal{L}_\text{align}}
\newcommand{\Lclean}{\mathcal{L}_\text{clean}}
\newcommand{\lhat}[1]{\hat{\ell}_{#1}}

\icmltitlerunning{KALE: Kernel Alignment with Loss Equilibration for Web-Scale CLIP--DINOv2 Alignment}

\begin{document}

\twocolumn[
\icmltitle{KALE: Kernel Alignment with Loss Equilibration\\
           for Stable CLIP--DINOv2 Alignment at Web Scale}

\begin{icmlauthorlist}
\icmlauthor{Michał Pawłowicz}{inst}
\end{icmlauthorlist}

\icmlaffiliation{inst}{TODO: Institution}
\icmlkeywords{CLIP, DINOv2, kernel alignment, web-scale training, dynamic loss weighting}

\vskip 0.3in
]

\begin{abstract}
Kernel-based alignment of CLIP toward a vision-centric teacher such as DINOv2
(KUEA) improves CLIP's visual representations while preserving text-encoder
compatibility, using a fixed trade-off weight tuned on curated ImageNet-1K\@.
We ask whether this transfers to noisy, web-scale data (CC12M) and find that it
does not: the alignment term's weighted contribution falls to about 0.2\% of the
clean term, so under any fixed weight its gradient is effectively inert.
We introduce KALE, a loss-equilibration controller that tracks both losses and
adaptively rescales the alignment weight toward a target ratio, restoring the
signal with no per-dataset tuning; reaching balance requires increasing the
weight by roughly four orders of magnitude, and the required value is
configuration-dependent, so no fixed scalar suffices.
We characterize the resulting regime: a bounded high learning rate and a decaying
schedule with a moderate floor are needed for stability, and the controller
equilibrates rather than diverging.
On a 3.3M-image CC12M subset, the aligned model preserves image--text retrieval
and reproducibly improves SVHN linear probing; zero-shot improves by +2.00 over
CLIP on the standard 11-dataset average, exceeding KUEA's +1.29.
We report all results with explicit run-to-run variance and base our conclusions
on the metrics that are stable across runs.
\end{abstract}

\section{Introduction}
\label{sec:intro}

Vision--language models such as CLIP \citep{CLIP} offer strong zero-shot
performance but can struggle on tasks requiring fine-grained visual
discrimination \citep{Tong2024}.
Recent work on kernel-based CLIP$\to$DINOv2 alignment \citep{KUEA} showed that
aligning the CLIP visual encoder towards a DINOv2 teacher \citep{DINOv2} improves
zero-shot accuracy and especially linear probing performance on curated
ImageNet-scale data, while preserving compatibility with CLIP's text encoder.
However, it remains unclear whether this alignment approach can be made stable
and effective at web-scale on noisy, heterogeneous data such as
CC12M \citep{CC12M}.

KUEA optimizes a combination of a kernel-alignment loss between student and
teacher embeddings and a ``clean'' loss that penalizes deviation from the
original CLIP features, balanced by a fixed scalar weight.
On curated ImageNet-1K, the authors report that performance is relatively
insensitive to this trade-off coefficient \citep{KUEA}.
At the scale of web-crawled data, we find that this apparent robustness does
not hold: the losses become highly imbalanced, the alignment signal can
effectively vanish, and na\"ively reusing a fixed weight leads to unstable or
ineffective training.

In this work we revisit kernel-based CLIP$\to$DINOv2 alignment at CC12M scale
and introduce a KALE controller that adaptively balances
the alignment and clean losses during training.
The controller increases alignment pressure when the alignment loss is too small
relative to the clean loss, and prevents it from dominating once the model has
moved sufficiently far from the original CLIP features.
We apply this controller to CLIP ViT-L/14, train on a 3.3M subset of CC12M, and
systematically study how learning rate, learning-rate schedule, and controller
parameters interact to determine stability and downstream performance.

KALE alignment improves over the original CLIP model, preserving image--text
retrieval, improving zero-shot classification, and reproducibly improving SVHN
linear probing.
Training on noisy CC12M rather than curated ImageNet yields a different downstream
operating point from KUEA: retrieval is preserved, SVHN linear probing is
improved, and the standard zero-shot average rises +2.00 over CLIP,
exceeding KUEA's +1.29 \citep{KUEA}.
The contribution is making
kernel-based alignment trainable at this scale without per-dataset weight tuning.
We also characterize the optimization regime: a high peak learning rate is needed
for the controller to track the alignment signal, training collapses if it is too
high, and a cosine schedule with a moderate floor (rather than a flat or fully
decayed one) keeps the representation's drift from CLIP bounded.

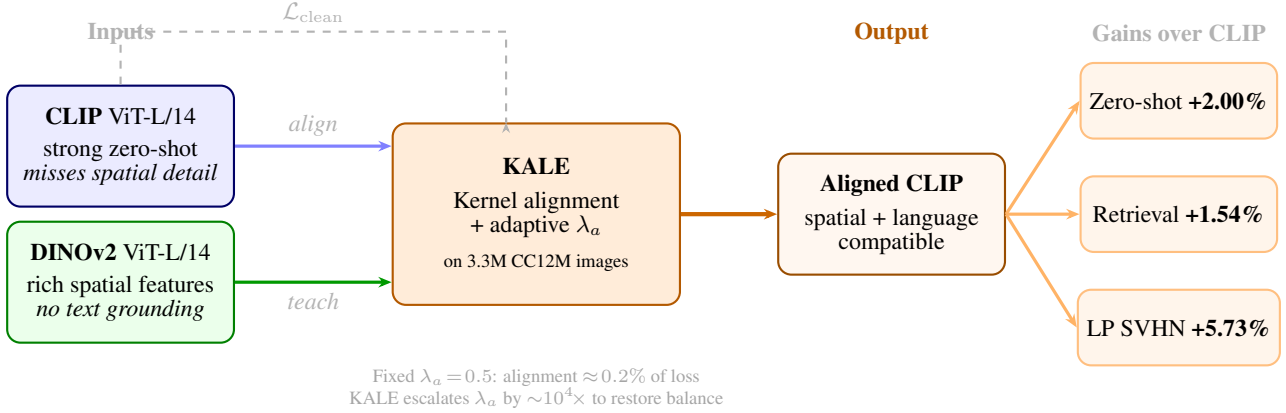
\begin{figure*}[t]
  \centering
\begin{tikzpicture}[
  font=\small,
  every node/.style={align=center},
  box/.style={draw, rounded corners=5pt, minimum width=30mm, minimum height=16mm,
              thick, align=center},
  clipbox/.style={box, fill=blue!8, draw=blue!40!black},
  dinobox/.style={box, fill=green!8, draw=green!50!black},
  kalebox/.style={box, fill=orange!15, draw=orange!70!black,
                  minimum width=38mm, minimum height=24mm},
  outbox/.style={box, fill=orange!6, draw=orange!60!black, minimum width=30mm},
  taskrow/.style={draw=none, fill=none},
  arr/.style={-{Stealth[length=5pt,width=4pt]}, line width=1.2pt},
  darr/.style={-{Stealth[length=4pt,width=3pt]}, dashed, line width=0.8pt, gray!60},
  lbl/.style={font=\footnotesize\itshape, text=gray!60},
  head/.style={font=\footnotesize\bfseries},
  note/.style={font=\scriptsize, text=gray!70, align=center},
]

\node[clipbox] (clip) at (0,0.9)
  {\textbf{CLIP} ViT-L/14\\[2pt]\footnotesize strong zero-shot\\[-1pt]\footnotesize \textit{misses spatial detail}};

\node[dinobox] (dino) at (0,-0.9)
  {\textbf{DINOv2} ViT-L/14\\[2pt]\footnotesize rich spatial features\\[-1pt]\footnotesize \textit{no text grounding}};

\node[head, text=gray!60] at (0, 2.4) {Inputs};

\node[kalebox] (kale) at (5.5, 0) {
  \textbf{KALE}\\[3pt]
  \footnotesize Kernel alignment\\[-1pt]
  \footnotesize + adaptive $\lambda_a$\\[3pt]
  \scriptsize on 3.3M CC12M images
};


\node[note] at (5.5, -2.3) {
  Fixed $\lambda_a\!=\!0.5$: alignment $\approx\!0.2\%$ of loss\\
  KALE escalates $\lambda_a$ by ${\sim}10^4\!\times$ to restore balance
};

\node[outbox] (aligned) at (10.2, 0) {
  \textbf{Aligned CLIP}\\[3pt]
  \footnotesize spatial + language\\[-1pt]
  \footnotesize compatible
};

\node[head, text=orange!70!black] at (10.2, 2.4) {Output};

\node[head, text=gray!60] at (14.0, 2.4) {Gains over CLIP};

\node[box, fill=orange!8, draw=orange!50, minimum width=26mm, minimum height=10mm]
  (t1) at (14.0, 1.5) {\footnotesize Zero-shot \textbf{+2.00\%}};
\node[box, fill=orange!8, draw=orange!50, minimum width=26mm, minimum height=10mm]
  (t2) at (14.0, 0)   {\footnotesize Retrieval \textbf{+1.54\%}};
\node[box, fill=orange!8, draw=orange!50, minimum width=26mm, minimum height=10mm]
  (t3) at (14.0,-1.5) {\footnotesize LP SVHN \textbf{+5.73\%}};

\draw[arr, blue!50] (clip.east) -- node[lbl, above]{align} (kale.west |- clip.east);
\draw[arr, green!60!black] (dino.east) -- node[lbl, below]{teach} (kale.west |- dino.east);
\draw[arr, orange!80!black, line width=1.6pt] (kale.east) -- (aligned.west);
\draw[arr, orange!60] (aligned.east) -- (t1.west);
\draw[arr, orange!60] (aligned.east) -- (t2.west);
\draw[arr, orange!60] (aligned.east) -- (t3.west);

\draw[darr] (clip.north) -- ++(0,0.7) -| node[near start, above, lbl]
  {$\mathcal{L}_\mathrm{clean}$} ([xshift=-4mm]kale.north) -- ([xshift=-4mm]kale.north);

\end{tikzpicture}
  \caption{%
    \textbf{Overview of this work.}
    We take a pre-trained CLIP visual encoder and align it toward a frozen
    DINOv2 teacher in kernel space, using a KALE
    controller that adaptively rescales the alignment loss weight on
    web-crawled CC12M data.
    A fixed weight ($\lambda_a{=}0.5$) is inert at this scale — the
    alignment term is ${\approx}0.2\%$ of the clean loss; the controller
    escalates $\lambda_a$ by up to $47{,}000\times$ to restore balance.
    The resulting encoder retains text compatibility and improves zero-shot,
    retrieval, and linear-probing performance (gains over CLIP shown;
    see Table~\ref{tab:main}).}
  \label{fig:concept}
\end{figure*}

\begin{figure*}[t]
  \centering
  \includegraphics[width=\linewidth]{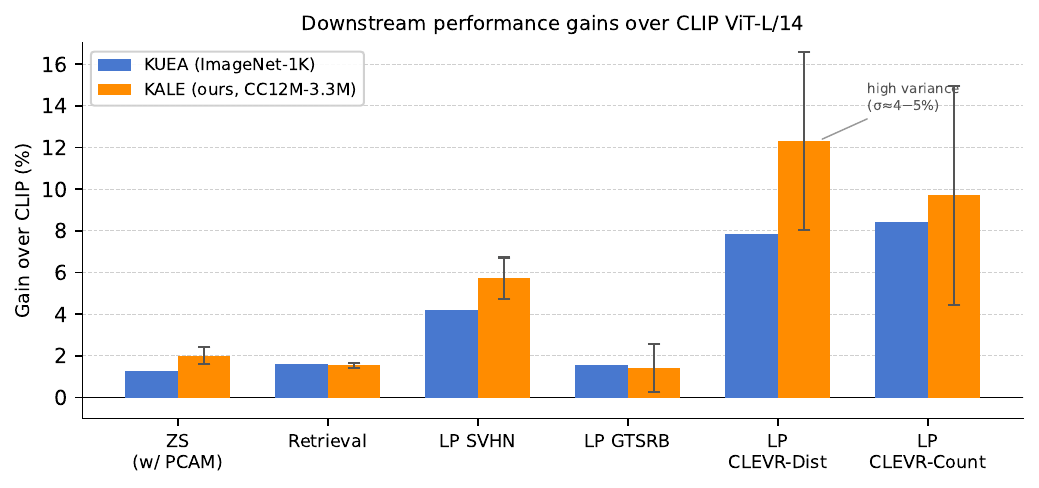}
  \caption{%
    \textbf{Overview of downstream gains over CLIP ViT-L/14.}
    Bars show the gain over CLIP in percentage points, grouped into zero-shot,
    retrieval, and linear-probing tasks.
    KUEA (blue) trains on curated ImageNet-1K with a fixed loss weight;
    our KALE (orange) trains on noisy CC12M-3.3M with the adaptive controller.
    Error bars on KALE show run-to-run standard deviation across seven
    replicates; CLEVR tasks are high-variance (\autoref{sec:limitations}).
    KALE preserves retrieval, matches or exceeds KUEA on SVHN, and improves the
    standard zero-shot average --- a distinct operating point reached without
    per-dataset weight tuning.}
  \label{fig:overview}
\end{figure*}

\par\medskip
\noindent\begin{minipage}{\linewidth}
\paragraph{Contributions.}
\begin{itemize}
  \item We extend kernel-based CLIP$\to$DINOv2 alignment from curated
    ImageNet-1K to noisy, web-scale CC12M training data by introducing a
    KALE controller that keeps the alignment loss
    influential throughout training.

  \item We conduct a detailed training-dynamics study of KALE alignment,
    showing that web-scale alignment requires a high peak learning rate within
    a bounded stability envelope and a decaying schedule with a moderate floor
    (not a flat one), and characterizing how the controller equilibrates.

  \item On a 3.3M subset of CC12M, the aligned model reaches a different
    downstream operating point from the ImageNet-trained KUEA checkpoint ---
    retrieval preserved, SVHN linear probing improved, and the standard
    zero-shot average improved over CLIP --- without per-dataset weight tuning,
    reported with an explicit run-to-run variance analysis.
\end{itemize}
\end{minipage}

\section{Related Work}
\label{sec:related}

\paragraph{Vision--language pretraining and CLIP finetuning.}
CLIP \citep{CLIP} trains joint image--text encoders on web-scale data and serves
as the backbone for a wide range of downstream vision and language tasks.
A large body of work adapts CLIP to specific tasks or domains: prompt tuning
\citep{CoOp} and adapter-based methods \citep{TipAdapter,CLIPAdapter} modify how
CLIP's frozen features are used without altering the encoder.
Recent work shows that CLIP's visual encoder has systematic fine-grained
discrimination failures \citep{Tong2024}, motivating direct improvements to the
visual representation.
KUEA \citep{KUEA} and our work take this route, modifying the visual encoder via
unsupervised alignment to a DINOv2 teacher while keeping the text encoder and
its language grounding intact.

\paragraph{Kernel-based CLIP--DINOv2 alignment.}
\citet{KUEA} propose aligning the CLIP visual encoder toward DINOv2 by minimizing
the discrepancy between their kernel similarity matrices over batches of images,
while penalizing deviation from the original CLIP embeddings via an L2
regularization term.
Trained on ImageNet-1K with a fixed trade-off weight between the two losses,
KUEA improves zero-shot classification and linear probing without breaking
compatibility with CLIP's text encoder.
Our work takes KUEA as its starting point and asks whether this alignment can be
made effective on noisy, web-crawled data at much larger scale --- a different
training regime with a larger batch, a higher learning rate, and substantially
noisier data.
We find that the fixed-weight formulation fails in this regime due to loss
imbalance, and address it with the dynamic penalty-weight controller of
\autoref{sec:kale}.

\paragraph{DINOv2 and dense visual features.}
DINOv2 \citep{DINOv2} produces dense visual features that perform strongly on
both global and local tasks, complementing CLIP's language grounding with
stronger spatial representations.
AM-RADIO \citep{AM-RADIO} distills CLIP, DINOv2, and SAM jointly into a single
student trained on DataComp-1B \citep{DataComp}, unifying their capabilities in
one model.
Our work is complementary and far more lightweight: rather than training a new
backbone, we adapt an existing CLIP ViT-L/14 checkpoint toward a DINOv2 teacher
on a 3.3M-sample CC12M subset, aiming to improve fine-grained visual
representations while preserving text compatibility.

\paragraph{Dynamic loss weighting.}
Balancing competing objectives in multi-loss training is a well-studied problem.
GradNorm \citep{GradNorm} reweights task losses by normalizing gradient magnitudes
to equalize task learning rates.
\citet{Kendall2018} learn loss weights as homoscedastic uncertainty parameters.
Dynamic Weight Averaging \citep{DWA} adjusts weights based on the relative rate
of change of each task's loss.
These approaches operate on gradient norms or introduce additional learned
parameters, and are designed for multi-task settings where tasks compete on
separate data.
Our KALE controller addresses a different problem: at CC12M scale the alignment
and clean losses are computed on the same data but differ in magnitude by orders
of magnitude, so their weighted ratio rather than their gradient balance is what
must be controlled.
The controller tracks exponential moving averages of both losses and adjusts a
single penalty weight every $\Delta$ steps to drive their ratio toward a target
$\tau$, with no additional parameters.
We quantify the imbalance and the controller's response in \autoref{sec:dynamics}.

\paragraph{Web-scale training and learning-rate schedule.}
Training on web-crawled data introduces noise and domain diversity far exceeding
curated datasets, which changes the optimization landscape for alignment
objectives.
Our secondary finding concerns the learning-rate schedule: at CC12M scale with a
dynamic controller, the schedule matters more than in standard fine-tuning.
A flat learning rate lets CLIP drift grow unchecked throughout training, degrading
downstream performance; a cosine schedule that decays to zero is also harmful, as
the late-stage step size becomes too small for the controller to maintain balance.
A decaying schedule with a moderate non-zero floor keeps drift bounded and the
dual-loss system stable.
We analyze the stability envelope --- including the collapse regime at high peak
learning rates --- in \autoref{sec:dynamics}.

\section{Method}
\label{sec:method}

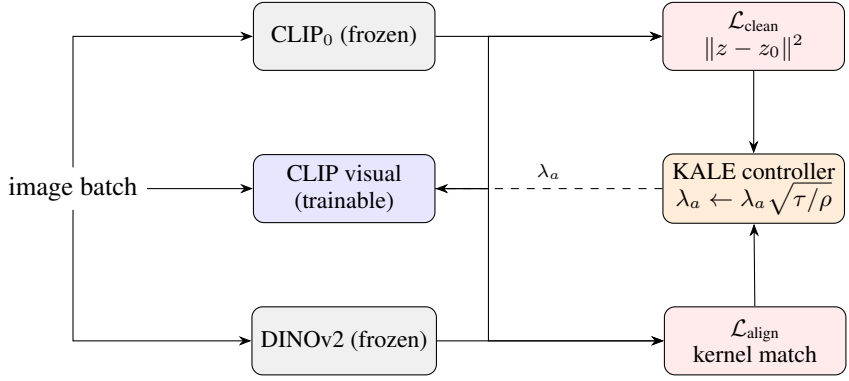
\begin{figure*}[t]
  \centering
\begin{tikzpicture}[node distance=11mm and 30mm,
  box/.style={draw,rounded corners,minimum height=9mm,minimum width=24mm,align=center,font=\small},
  frozen/.style={box,fill=gray!12}, train/.style={box,fill=blue!10},
  loss/.style={box,fill=red!8}, ctl/.style={box,fill=orange!15},
  >={Stealth[]}]
  \node[train] (clip) {CLIP visual\\(trainable)};
  \node[frozen, above=of clip] (clip0) {CLIP$_0$ (frozen)};
  \node[frozen, below=of clip] (dino)  {DINOv2 (frozen)};
  \node[left=14mm of clip] (img) {image batch};
  \node[loss, right=of clip0] (clean) {$\mathcal{L}_\text{clean}$\\$\lVert z - z_0\rVert^2$};
  \node[ctl,  right=of clip]  (ctrl)  {KALE controller\\$\lambda_a \leftarrow \lambda_a\sqrt{\tau/\rho}$};
  \node[loss, right=of dino]  (align) {$\mathcal{L}_\text{align}$\\kernel match};
  \draw[->] (img) -- (clip);
  \draw[->] (img.north) |- (clip0.west);
  \draw[->] (img.south) |- (dino.west);
  \draw[->] (clip0) -- (clean);
  \draw[->] (dino)  -- (align);
  \draw[->] (clip.east) -- ++(7mm,0) |- (clean.west);
  \draw[->] (clip.east) -- ++(7mm,0) |- (align.west);
  \draw[->] (clean) -- (ctrl);
  \draw[->] (align) -- (ctrl);
  \draw[->,dashed] (ctrl.west) -- (clip.east)
    node[midway,above,font=\scriptsize]{$\lambda_a$};
\end{tikzpicture}
  \caption{%
    KALE alignment setup: trainable CLIP student, frozen DINOv2 teacher and
    frozen CLIP$_0$ reference, the kernel-alignment and clean losses, and the
    KALE controller updating~$\la$.}
  \label{fig:method}
\end{figure*}

Figure~\ref{fig:method} summarises the setup: a trainable CLIP visual encoder
aligned to a frozen DINOv2 teacher in kernel space, regularised toward the frozen
original CLIP encoder by a clean loss, with the KALE controller setting the
alignment weight $\la$ from the running loss ratio.

\subsection{Kernel-based CLIP$\to$DINOv2 alignment (KUEA recap)}
\label{sec:kuea}

We build on the kernel-based unsupervised embedding alignment (KUEA
\citep{KUEA}) framework, which aligns the CLIP visual encoder towards a DINOv2
visual teacher by matching their similarity structure over batches of images.
Given a batch of images, KUEA computes student embeddings using CLIP's visual
encoder and teacher embeddings using DINOv2, forms kernel similarity matrices in
each embedding space, and minimises a discrepancy between these matrices.
In parallel, it penalises the deviation of the student embeddings from the
original CLIP embeddings via a ``clean'' loss.
The total objective is a weighted sum of the kernel-alignment loss and the clean
loss, controlled by a fixed scalar weight that trades off alignment against
preserving CLIP's original visual--text grounding.

\paragraph{Notation.}
Let $f_\theta$ denote the CLIP visual encoder being fine-tuned, $f_0$ the frozen
original CLIP encoder, and $g$ the frozen DINOv2 teacher.
For a batch of $n$ images $\{x_i\}_{i=1}^n$, let $z_i = f_\theta(x_i)$,
$z_i^0 = f_0(x_i)$, and $v_i = g(x_i)$ denote the student, original CLIP, and
teacher embeddings respectively.

\paragraph{Kernel matrices.}
KUEA matches the similarity structure of the student and teacher embedding spaces
via kernel matrices.
We use a normalised polynomial kernel of degree~3.
For a set of embeddings $\{u_i\}$, the $(i,j)$ entry of the kernel matrix is
\begin{equation}
  A_{ij}(u) = \Bigl(\tfrac{u_i^\top u_j}{d} + 1\Bigr)^3,
  \quad
  K_{ij}(u) = \frac{A_{ij}(u)}{\sqrt{A_{ii}(u)\cdot A_{jj}(u)}},
  \label{eq:kernel}
\end{equation}
where $A$ is the unnormalised polynomial kernel and $d$ is the embedding
dimension.
This normalisation ensures $K_{ii}=1$ and the matrix is scale-invariant.

\paragraph{Kernel alignment loss.}
The alignment loss minimises the mean squared difference between the student and
teacher kernel matrices over the batch:
\begin{equation}
  \Lalign(\theta) = \frac{1}{n^2}\sum_{i,j}
    \bigl(K_{ij}(z) - K_{ij}(v)\bigr)^2.
  \label{eq:Lalign}
\end{equation}

\paragraph{Clean loss.}
To prevent the student from drifting too far from the original CLIP
representations (which would degrade zero-shot and retrieval performance),
KUEA adds a clean loss penalising the L2 distance between student and original
CLIP embeddings:
\begin{equation}
  \Lclean(\theta) = \frac{1}{n}\sum_i \|z_i - z_i^0\|^2.
  \label{eq:Lclean}
\end{equation}

\paragraph{Total objective.}
The full training objective is a weighted combination:
\begin{equation}
  \mathcal{L}(\theta) = \lc\cdot\Lclean(\theta) + \la\cdot\Lalign(\theta),
  \label{eq:total}
\end{equation}
where $\lc$ is the clean weight (fixed at 1.0 in our experiments) and $\la$ is
the penalty weight controlling alignment strength.
In the original KUEA, $\la$ is a fixed scalar; in our work, we replace it with
the KALE controller described in \autoref{sec:kale}.

\subsection{KALE controller}
\label{sec:kale}

At CC12M scale the kernel-alignment loss is much smaller than the clean loss, so
its gradient effectively vanishes under a fixed scalar weight (quantified in
\autoref{sec:dynamics}); na\"ively raising the weight instead lets the alignment
term dominate late in training and push the model too far from CLIP's original
representations.
We replace the fixed trade-off weight with a KALE
controller that adjusts alignment strength during training.

The controller updates the penalty weight from the running ratio of the two loss
contributions: it raises the weight when the alignment loss is too small relative
to the clean loss, and stops raising it (with an optional cap) once the alignment term is large enough
to risk overwhelming the CLIP-consistency signal.

\paragraph{EMA tracking.}
The controller maintains exponential moving averages of both losses to smooth out
per-batch noise:
\begin{align}
  \lhat{align}^{(t)} &= \alpha\,\lhat{align}^{(t-1)}
                      + (1-\alpha)\,\Lalign^{(t)}, \nonumber\\
  \lhat{clean}^{(t)} &= \alpha\,\lhat{clean}^{(t-1)}
                      + (1-\alpha)\,\Lclean^{(t)},
  \label{eq:ema}
\end{align}
with decay $\alpha=0.99$; the EMAs are initialised to the first batch's loss
values.

\paragraph{Loss ratio.}
At each update step (every $\Delta$ training steps, with $\Delta=100$ in our
experiments), the controller computes the current ratio of the alignment
contribution to the clean contribution:
\begin{equation}
  \rho^{(t)} = \frac{\la^{(t)}\cdot\lhat{align}^{(t)}}
                    {\lc\cdot\lhat{clean}^{(t)}}.
  \label{eq:ratio}
\end{equation}

\paragraph{Weight update.}
The penalty weight is updated to drive $\rho$ toward a target value $\tau$:
\begin{equation}
  \la^{(t+1)} = \la^{(t)}\cdot\sqrt{\frac{\tau}{\rho^{(t)}}}.
  \label{eq:update}
\end{equation}
The square-root dampening prevents overcorrection: if the alignment contribution
is half the target, the weight is multiplied by $\sqrt{2}$ rather than~2, giving
the optimiser time to adapt before the next update.
The controller is inactive during warmup steps.

\begin{algorithm}[t]
   \caption{KALE training (one configuration)}
   \label{alg:kale}
\begin{algorithmic}
   \STATE {\bfseries Input:} student $f_\theta$, frozen teacher $g$, frozen reference $f_0$, data $\mathcal{D}$
   \STATE {\bfseries Params:} clean weight $\lc$, initial $\la$, target $\tau$, interval $\Delta$, EMA decay $\alpha$, warmup $W$
   \STATE Initialise EMAs $\lhat{align},\lhat{clean}$ to the first batch's losses
   \FOR{$t=1$ {\bfseries to} $T$}
   \STATE sample batch; compute $\Lalign,\Lclean$ \hfill (Eq.~\ref{eq:Lalign},~\ref{eq:Lclean})
   \STATE update $f_\theta$ on $\lc\Lclean+\la\Lalign$ \hfill (Eq.~\ref{eq:total})
   \STATE $\lhat{align}\leftarrow\alpha\,\lhat{align}+(1-\alpha)\,\Lalign$ \hfill (Eq.~\ref{eq:ema})
   \STATE $\lhat{clean}\leftarrow\alpha\,\lhat{clean}+(1-\alpha)\,\Lclean$
   \IF{$t>W$ {\bfseries and} $t \bmod \Delta = 0$}
   \STATE $\rho \leftarrow (\la\,\lhat{align})/(\lc\,\lhat{clean})$ \hfill (Eq.~\ref{eq:ratio})
   \STATE $\la \leftarrow \la\,\sqrt{\tau/\rho}$ \hfill (Eq.~\ref{eq:update})
   \ENDIF
   \ENDFOR
   \STATE {\bfseries return} $f_\theta$
\end{algorithmic}
\end{algorithm}

\paragraph{Intuition.}
When the alignment loss is small relative to the clean loss ($\rho\ll\tau$), the
controller increases $\la$ to amplify the alignment signal.
When the alignment loss dominates ($\rho\gg\tau$), the controller reduces $\la$
to protect the CLIP-consistency signal.
This creates a feedback loop that keeps both losses contributing meaningfully
throughout training, without requiring manual tuning of the trade-off coefficient
for each new dataset or scale.

\begin{figure*}[t]
  \centering
  \includegraphics[width=\linewidth]{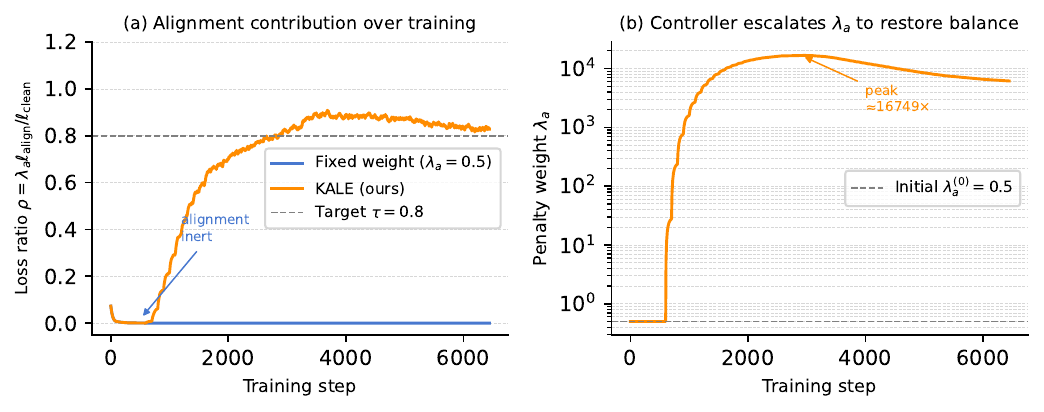}
  \caption{%
    \textbf{Why KALE is needed.}
    \textit{(a)} Under a fixed weight ($\la{=}0.5$, blue), the loss ratio
    $\rho$ collapses to near zero within the first steps and stays there ---
    the alignment gradient is effectively inert throughout training.
    The KALE controller (orange) drives $\rho$ to the target $\tau{=}0.8$
    and maintains it.
    \textit{(b)} To achieve this, the controller escalates $\la$ by
    ${\sim}10^4{\times}$ before it equilibrates (\autoref{sec:dynamics}).
    The counterfactual fixed-weight $\rho$ is derived analytically from the
    logged KALE run (see \autoref{sec:dynamics}).}
  \label{fig:dpw_intuition}
\end{figure*}

\paragraph{Fixed point and stability.}
The update admits a simple analysis under timescale separation: because the
controller fires only every $\Delta$ steps on EMA-smoothed losses
($\alpha=0.99$), the losses are approximately constant between updates.
Writing $\kappa = \lhat{align}/(\lc\,\lhat{clean})$ for the (quasi-static)
unweighted loss ratio, the effective ratio is
$\rho^{(t)}=\kappa\,\la^{(t)}$ and the update becomes
\begin{equation}
  \la^{(t+1)}
    = \la^{(t)}\sqrt{\frac{\tau}{\kappa\,\la^{(t)}}}
    = \sqrt{\frac{\tau}{\kappa}}\,\bigl(\la^{(t)}\bigr)^{1/2}.
  \label{eq:update-simplified}
\end{equation}
In log-weight coordinates $u=\log\la$ this is the linear recurrence
$u^{(t+1)}=\tfrac{1}{2}u^{(t)}+\tfrac{1}{2}\log(\tau/\kappa)$,
with a unique fixed point
\begin{equation}
  \la^\star = \frac{\tau}{\kappa}
            = \tau\,\frac{\lc\,\lhat{clean}}{\lhat{align}}, \qquad
  \rho^\star = \tau.
  \label{eq:fixedpoint}
\end{equation}
The controller's equilibrium sets the alignment contribution to exactly $\tau$
times the clean contribution --- its design target.
Deviations contract as $u^{(t+1)}-u^\star = \tfrac{1}{2}(u^{(t)}-u^\star)$,
so $\la$ converges geometrically to $\la^\star$ from any positive initialisation,
at rate $\tfrac{1}{2}$ per update and without oscillation.
The square-root exponent sets this rate: with a general exponent $p$ in
$\la^{(t+1)}=\la^{(t)}(\tau/\rho^{(t)})^p$ the contraction factor is $|1-p|$,
so $p=1$ converges in a single step under static losses but overshoots once the
losses drift or the ratio estimate is noisy, and $p>2$ diverges ($p=2$ is
marginally stable, oscillating);
$p=\tfrac{1}{2}$ trades one-step convergence for a monotone, robust approach
under the moving, EMA-smoothed losses.

This analysis is local: it treats the losses as fixed between updates.
In practice $\lhat{align}$ falls as the student aligns and $\lhat{clean}$ first
rises (as the representation drifts from CLIP) and later recedes, so $\la^\star$
is a moving target the controller tracks rather than a constant.
The contraction is consistent with the equilibration observed in
\autoref{sec:dynamics}; it does not characterise the fully coupled system of
controller, optimiser, and learning-rate schedule, whose stability envelope (the
collapse at high learning rate, \autoref{sec:dynamics}) lies outside this
fixed-point argument.

\subsection{Training at web scale on CC12M}
\label{sec:webscale}

Unlike KUEA \citep{KUEA}, which trains on curated ImageNet-1K, we train on noisy
web-crawled CC12M \citep{CC12M} (images only, with CLIP's text encoder frozen so
zero-shot capability is preserved).
It is at this scale --- with far greater domain variation than the ImageNet label
space --- that the fixed trade-off weight proves insufficient, motivating the
controller of \autoref{sec:kale}.

\section{Experiments}
\label{sec:experiments}

\subsection{Experimental setup}
\label{sec:setup}

We evaluate our approach by fine-tuning the CLIP ViT-L/14 (openai) visual
encoder towards a DINOv2 teacher using our KALE-augmented kernel-alignment
objective. The teacher is a DINOv2 model with dense visual features, and the
CLIP text encoder remains frozen so that zero-shot prompting remains valid.
Training is performed on a 3.3M random subset of CC12M, using large-batch
distributed training. As baselines, we compare against the original CLIP
ViT-L/14 model and the publicly released KUEA checkpoint trained on ImageNet-1K.

\paragraph{Models.}
We fine-tune the CLIP ViT-L/14 visual encoder (OpenAI pretrained) using
DINOv2 ViT-L/14 with registers \citep{Darcet2024,DINOv2} as the frozen teacher.
The CLIP text encoder is kept frozen throughout.
Both student and teacher use a normalised polynomial kernel of degree~3,
consistent with KUEA \citep{KUEA}.
The CLIP kernel parameters ($\gamma$, $c_0$) are initialised to values
pre-computed by minimising the L2 distance between CLIP and DINOv2 kernel
matrices on a small sample of ImageNet images ($\gamma=0.0032$,
$c_0=0.1916$) and treated as learnable during fine-tuning.

\paragraph{Training data.}
We train on a 3.3M random subset of CC12M \citep{CC12M}, using images only.
To avoid loading three large models simultaneously during training, we
precompute and cache the CLIP and DINOv2 embeddings for all 3.3M images before
training.
The same fixed subset and precomputed embeddings are used across all experiments,
ensuring that differences in results reflect hyperparameter choices rather than
data sampling.

\paragraph{Training setup.}
We train with AdamW \citep{AdamW} ($\beta_1=0.9$, $\beta_2=0.999$, weight
decay $10^{-4}$) using distributed data parallel across $8{\times}$ H100 80\,GB
GPUs (AWS p5.48xlarge).
Per-GPU batch size is 128, giving an effective batch size of 1024.
The learning rate follows a cosine decay schedule with a 10\% linear warmup and
a minimum LR floor of 10\% of peak LR by default (varied in \autoref{sec:ablation_lr}).
Training runs for 2 epochs unless otherwise noted.
All experiments use bfloat16 mixed precision.

\paragraph{KALE configuration.}
For runs with the dynamic penalty-weight controller, we set the update interval
$\Delta=100$ steps, target ratio $\tau=0.8$, EMA decay $\alpha=0.99$, and
initial penalty weight $\la^{(0)}=0.5$.
The clean weight is fixed at $\lc=1.0$ throughout.

\paragraph{Baselines.}
We compare against (1) the original CLIP ViT-L/14 (OpenAI), and (2) KUEA
\citep{KUEA} --- CLIP ViT-L/14 aligned on ImageNet-1K with the original fixed
trade-off weight.
For zero-shot and linear-probing comparisons we use KUEA's published results
\citep[Tables~1 and~3]{KUEA}; our evaluation pipeline reproduces their published
CLIP baseline exactly on every zero-shot and linear-probing dataset (zero-shot
average 65.26, linear-probing average 50.59), so gains over CLIP are directly
comparable across the two works.
For retrieval we instead report KUEA reproduced in our identical pipeline, because
the paper reports per-recall (R@k) breakdowns under a retrieval protocol that our
pipeline does not match, leaving no single comparable published figure.

\paragraph{Evaluation.}
We follow the CLIP\_benchmark \citep{CLIPBenchmark} evaluation suite used in
KUEA\@.
For zero-shot classification we evaluate on 11 datasets:
CIFAR-10/100 \citep{CIFAR},
Caltech-101 \citep{Caltech101},
OxfordPets \citep{OxfordPets},
DTD \citep{DTD},
FER2013 \citep{FER2013},
PCAM \citep{PCAM},
RESISC45 \citep{RESISC45},
EuroSAT \citep{EuroSAT},
ImageNet-Sketch \citep{ImageNetSketch},
and ImageNet-O \citep{ImageNetO};
ImageNet-1K itself is used only as an in-training selection signal
(\autoref{sec:dynamics}), not in the reported average \citep{ImageNet}.
Table~\ref{tab:main} reports the standard zero-shot average over these datasets
and the mean linear-probing gain.
For linear probing we evaluate on CLEVR Distance and CLEVR Counts \citep{CLEVR},
GTSRB \citep{GTSRB}, and SVHN \citep{SVHN}.
For image--text retrieval we report Recall@1/5/10 on Flickr30K \citep{Flickr30K}
and MSCOCO \citep{MSCOCO}.

\paragraph{Checkpoint selection.}
For every run we report the checkpoint chosen by a fixed \emph{controller-convergence}
rule: the first checkpoint at which the EMA-smoothed loss ratio $\rho$ reaches the
target $\tau$ (the headline run converges at $\approx$ step~4025, about 60\% through
the two-epoch budget), subject to a validity filter that rejects collapsed states
(CLIP drift $\le 8$, embedding effective rank $\ge 90$).
Here CLIP drift is the mean L2 distance of the student embeddings from the frozen
CLIP reference, and the embedding effective rank is $\exp$ of the Shannon entropy
of the batch embeddings' normalised singular-value spectrum
\citep{EffectiveRank} --- a continuous count
of active dimensions that drops sharply under collapse.
The rule uses only training-side signals, with ImageNet-1K as an in-training
monitor; it is applied identically to all runs, so no per-run test-set tuning
enters checkpoint choice.

\paragraph{Run-to-run variance.}
We quantify variance over seven replicates of the recommended configuration that
differ only in the minimum learning-rate floor (0--25\% of peak).
Cosine decay sits above all these floors until $\approx$ step~4300 --- after the
reported checkpoint (step~4025) --- and we verified the learning rate is identical
($3.57\times10^{-5}$) across all seven at that step.
Data order is fixed (deterministic sampler, seed~0), so the only uncontrolled
source of variation through the checkpoint is GPU nondeterminism.
We therefore treat the spread as run-to-run noise of a single configuration, not
a hyperparameter sweep.

\subsection{Main results}
\label{sec:results}

\begin{figure*}[t]
  \centering
  \includegraphics[width=\linewidth]{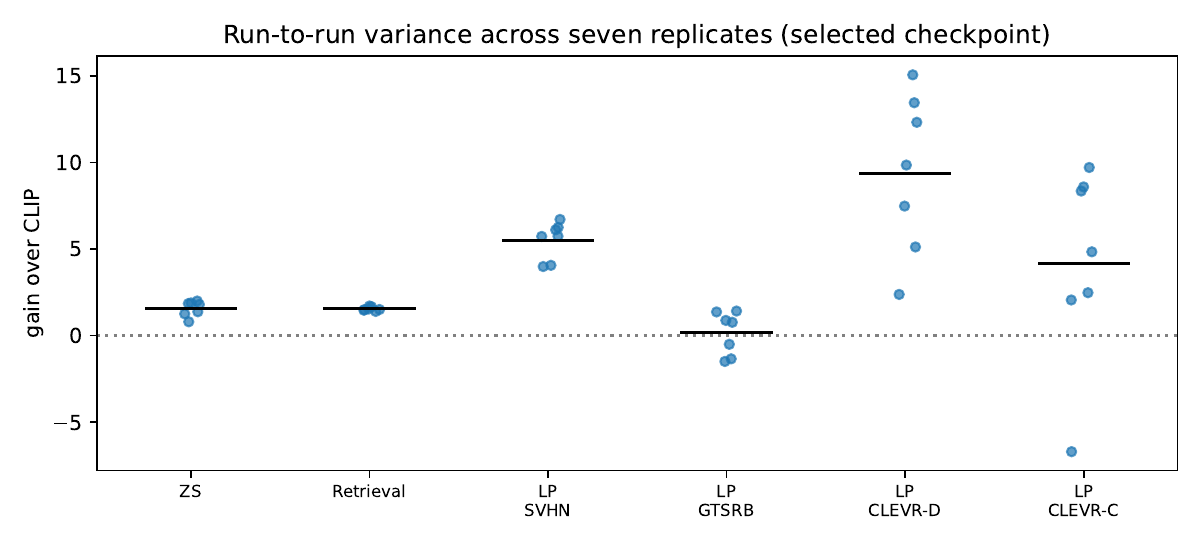}
  \caption{%
    Run-to-run variance across the seven identically-configured replicates
    (gain over CLIP at the selected checkpoint; bars mark the mean, points the
    individual runs).
    Zero-shot and retrieval are tightly clustered, whereas the CLEVR
    linear-probing tasks scatter by 13--16 points --- motivating the
    run-to-run $\sigma$ reported alongside the best run and the conclusions of
    \autoref{sec:limitations}.}
  \label{fig:variance}
\end{figure*}

We compare KALE-aligned CLIP (trained on CC12M-3.3M) against the original CLIP
ViT-L/14 and the KUEA checkpoint (trained on ImageNet-1K), using the identical
evaluation pipeline (\autoref{sec:setup}).
We report KALE as our \textbf{best single run} --- the recommended configuration
(learning rate $1{\times}10^{-4}$, $\tau=0.8$, two epochs), with the checkpoint
chosen by the controller-convergence rule of \autoref{sec:dynamics} --- matching
KUEA's single-run reporting.
Because linear probing exhibits substantial run-to-run variance, we report
alongside it the \textbf{run-to-run standard deviation across seven replicates
of the recommended configuration} (§\ref{sec:setup}); the spread reflects
training nondeterminism, not a hyperparameter sweep
(Figure~\ref{fig:variance}).
Zero-shot is reported on the standard 11-dataset average
(Table~\ref{tab:main}).

\begin{table*}[t]
  \centering
  \small
  \caption{%
    Downstream gains over CLIP ViT-L/14 (higher is better); \textbf{bold} marks
    the better result in each column.
    KUEA zero-shot and linear-probing values are the published results
    \citep[Tables~1 and~3]{KUEA}; KUEA retrieval ($\dagger$) is reproduced in
    our identical pipeline (\autoref{sec:setup}).
    The zero-shot column averages the same 11 datasets for both methods
    (ImageNet-1K excluded, used only as an in-training selection signal);
    LP is the mean linear-probing gain over the four LP datasets
    (SVHN, GTSRB, CLEVR-Dist, CLEVR-Count; per dataset in \autoref{tab:perdataset}).
    KALE is our best single run (recommended configuration; checkpoint by
    \autoref{sec:dynamics}); the $\sigma$ row is the run-to-run standard
    deviation across seven identically-configured replicates.}
  \label{tab:main}
  \begin{tabular}{lccc}
    \toprule
    \rowcolor{tableheader}
    Model (train data) & ZS & Ret & LP \\
    \midrule
    \addlinespace[2pt]
    KUEA (ImageNet-1K)         & $+1.29$ & $\mathbf{+1.63}^\dagger$ & $+5.51$ \\
    KALE (CC12M-3.3M), best run & $\mathbf{+2.00}$ & $+1.54$ & $\mathbf{+7.30}$ \\
    \quad run-to-run $\sigma$  & $0.41$  & $0.10$ & $2.03$ \\
    \bottomrule
  \end{tabular}
  \par\smallskip
  \footnotesize
  ZS\,=\,zero-shot (standard 11-dataset average, KUEA-comparable);
  Ret\,=\,retrieval;
  LP\,=\,mean linear-probing gain over SVHN, GTSRB, CLEVR-Dist, CLEVR-Count
  (per dataset in \autoref{tab:perdataset}). CLIP retrieval reference (absolute): 55.5.
  $\dagger$\,KUEA retrieval reproduced in our pipeline; paper reports R@k under
  a different protocol.
\end{table*}

\paragraph{Zero-shot and retrieval.}
These are the stable metrics: across the seven replicates retrieval varies by
$\sigma=0.10$ and zero-shot by $\sigma=0.41$, so the single-run numbers are
representative.
On retrieval, KALE trained on noisy web data essentially matches KUEA's
curated-ImageNet result (+1.54 vs.\ KUEA's +1.63 \citep{KUEA}).
On the standard 11-dataset average, zero-shot improves by +2.00 over CLIP,
exceeding KUEA's +1.29 \citep{KUEA}.
We base our conclusions on retrieval and zero-shot, the metrics that are stable
across runs.

\paragraph{Linear probing.}
Our best run improves SVHN by +5.73, above KUEA's published +4.19
\citep{KUEA}; this gain is reproducible --- all seven replicates land between
+4.0 and +6.7 ($\sigma=0.99$).
On the CLEVR tasks the best run also exceeds KUEA
(+12.32 vs.\ +7.85 on distance; +9.71 vs.\ +8.42 on counting)
\citep{KUEA}, but these tasks carry large run-to-run variance ($\sigma=4.3$
and~5.3).
GTSRB is within run-to-run noise (+1.42, $\sigma=1.16$).

\paragraph{Summary.}
The kernel-alignment objective, made trainable at web scale by the
controller (\autoref{sec:dynamics}), reaches a distinct operating point on noisy
CC12M without per-dataset weight tuning: it preserves retrieval
(+1.54 vs.\ +1.63), improves the standard zero-shot average (+2.00 vs.\ +1.29),
and improves SVHN linear probing (+5.73 vs.\ +4.19), with the structured CLEVR
tasks dominated by run-to-run variance.

Table~\ref{tab:perdataset} breaks the comparison down per dataset. On zero-shot,
our best run exceeds KUEA on the 11-dataset average and on Caltech-101,
OxfordPets, ImageNet-O, and PCAM; it trails on several of the remaining datasets
(e.g., CIFAR-10/100, RESISC45) and regresses on EuroSAT, which falls below CLIP.
On linear probing it exceeds KUEA on SVHN and both CLEVR tasks and is level on
GTSRB. The run-to-run spread behind these single-run numbers is shown in
Figure~\ref{fig:variance}, with the CLEVR tasks especially high-variance
(\autoref{sec:limitations}).

\begin{table*}[t]
  \centering
  \small
  \caption{%
    Per-dataset accuracy (\%) for CLIP, KUEA, and our best KALE run at the
    controller-selected checkpoint; \textbf{bold} marks the higher of KUEA and
    KALE in each row.
    Zero-shot and linear-probing values for CLIP and KUEA are the published
    numbers (\autoref{sec:setup}).
    Run-to-run spread across the seven replicates is shown in
    Figure~\ref{fig:variance}.}
  \label{tab:perdataset}
  \begin{tabular}{llrrr}
    \toprule
    \rowcolor{tableheader}
    Family & Dataset & CLIP & KUEA & KALE \\
    \midrule
    ZS & CIFAR-10        & 95.20 & $\mathbf{96.27}$ & 95.25 \\
    ZS & CIFAR-100       & 71.08 & $\mathbf{77.04}$ & 75.54 \\
    ZS & Caltech-101     & 83.30 & 84.32 & $\mathbf{85.21}$ \\
    ZS & FER2013         & 50.00 & $\mathbf{50.31}$ & 49.87 \\
    ZS & OxfordPets      & 93.21 & 93.40 & $\mathbf{93.59}$ \\
    ZS & DTD             & 55.21 & $\mathbf{55.74}$ & 55.69 \\
    ZS & RESISC45        & 63.35 & $\mathbf{63.83}$ & 63.48 \\
    ZS & EuroSAT         & 62.65 & $\mathbf{63.83}$ & 59.52 \\
    ZS & PCAM            & 52.00 & 52.68 & $\mathbf{67.12}$ \\
    ZS & ImageNet-Sketch & 59.59 & $\mathbf{59.67}$ & 59.49 \\
    ZS & ImageNet-O      & 32.25 & 34.90 & $\mathbf{35.05}$ \\
    \addlinespace
    ZS & Average         & 65.26 & 66.54 & $\mathbf{67.26}$ \\
    \midrule
    LP & SVHN            & 65.20 & 69.39 & $\mathbf{70.93}$ \\
    LP & GTSRB           & 72.94 & $\mathbf{74.51}$ & 74.36 \\
    LP & CLEVR-Dist      & 22.97 & 30.82 & $\mathbf{35.29}$ \\
    LP & CLEVR-Count     & 41.25 & 49.67 & $\mathbf{50.96}$ \\
    \bottomrule
  \end{tabular}
\end{table*}

\subsection{Learning rate and schedule ablations}
\label{sec:ablation_lr}

In this subsection we study how learning-rate magnitude and schedule affect the
stability of KALE alignment.
A high peak learning rate is needed for the controller to track the alignment
signal, but there is a sharp upper bound --- at $2{\times}10^{-4}$ training
collapses --- and a decaying schedule with a moderate floor, rather than a flat
one, keeps the CLIP drift bounded (\autoref{sec:dynamics}).
We frame these as stability findings; per-dataset downstream rankings across
learning rates fall within the run-to-run variance of \autoref{sec:limitations}.

\begin{table*}[t]
  \centering
  \small
  \caption{%
    Gains over CLIP and peak CLIP drift across learning rates
    (single run per LR; $1{\times}10^{-4}$ is the headline configuration).
    Training is stable up to $1{\times}10^{-4}$ and collapses at
    $2{\times}10^{-4}$ (drift runaway).
    Peak drift is the maximum L2 distance of the student embeddings from the
    frozen CLIP reference during training.
    Zero-shot and retrieval vary modestly across stable learning rates;
    LP is the mean gain over the four LP datasets (\autoref{tab:perdataset}).}
  \label{tab:lr}
  \begin{tabular}{lrrrrl}
    \toprule
    \rowcolor{tableheader}
    LR & ZS & Retrieval & LP & peak drift & status \\
    \midrule
    $4{\times}10^{-5}$ & $+1.42$ & $+1.09$ & $+5.02$ & 2.9  & stable \\
    $8{\times}10^{-5}$ & $+1.32$ & $+1.64$ & $+5.38$ & 5.1  & stable \\
    $1{\times}10^{-4}$ & $+2.00$ & $+1.54$ & $+7.30$ & 5.3  & stable \\
    $2{\times}10^{-4}$ & collapsed & collapsed & collapsed & 18.1 & collapse \\
    \bottomrule
  \end{tabular}
\end{table*}

\begin{table}[t]
  \centering
  \small
  \caption{%
    Learning-rate floor vs.\ end-of-training CLIP drift (L2) at the headline peak
    LR ($1{\times}10^{-4}$); single run per floor. As the floor rises and the
    schedule flattens, end-of-training drift grows monotonically across this
    single-run sweep --- a correlation we report as such, not a variance-bounded
    effect; moderate floors (0--25\%) differ by less than the run-to-run noise of
    \autoref{sec:limitations}. The embedding effective rank --- a continuous
    measure of how many embedding dimensions carry variance, with low values
    signalling collapse --- stays healthy ($\approx107$) for every floor, so none
    collapse: collapse arises only from excessive \emph{peak} LR
    (Table~\ref{tab:lr}), not from the floor.
    We omit transient peak drift here, as it is dominated by run-to-run
    nondeterministic spikes rather than the floor.}
  \label{tab:floor}
  \begin{tabular}{lcc}
    \toprule
    \rowcolor{tableheader}
    LR floor (\% of peak) & end-of-training drift & eff.\ rank \\
    \midrule
    0 (cosine $\to$ 0)      & 2.9      & 107 \\
    10--25 (moderate)       & 2.9--3.3 & 107 \\
    50                      & 3.9      & 107 \\
    75                      & 4.9      & 107 \\
    100 (flat)              & 6.2      & 107 \\
    \bottomrule
  \end{tabular}
\end{table}

\subsection{KALE and clean-weight ablations}
\label{sec:ablation_dpw}

Here we ablate the KALE target ratio and the clean-loss weight.
Increasing the clean-loss weight does not improve retrieval, indicating that the
linear-probing/retrieval balance is not controllable through that term alone.
We interpret the controller and clean-weight effects on linear probing
cautiously, as the per-configuration differences are comparable to the
run-to-run variance reported in \autoref{sec:limitations}.

\begin{table}[t]
  \centering
  \small
  \caption{%
    KALE target ratio and clean weight (gains over CLIP, headline learning rate).
    Raising the clean weight to 1.5 does not improve retrieval; the zero-shot and
    LP differences are within the run-to-run variance of
    \autoref{sec:limitations}.}
  \label{tab:kale}
  \begin{tabular}{lrrr}
    \toprule
    \rowcolor{tableheader}
    Setting & ZS & Retrieval & LP \\
    \midrule
    $\tau=0.8$ (default)   & $+2.00$ & $+1.54$ & $+7.30$ \\
    $\tau=1.0$             & $+1.05$ & $+1.77$ & $+7.23$ \\
    clean weight 1.5       & $+0.97$ & $+1.21$ & $+6.74$ \\
    \bottomrule
  \end{tabular}
\end{table}

\subsection{Training duration and checkpoint sensitivity}
\label{sec:ablation_dur}

\begin{figure}[t]
  \centering
  \includegraphics[width=0.75\linewidth]{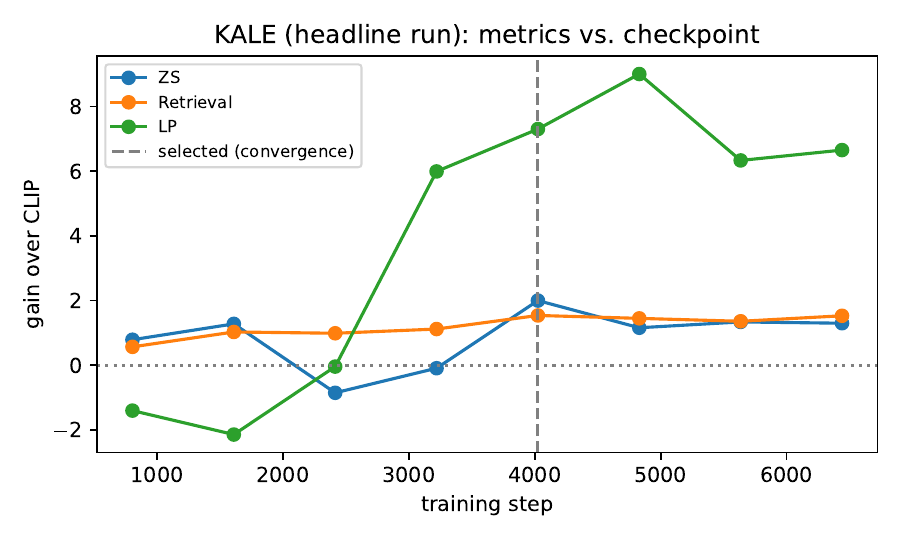}
  \caption{%
    Gains over CLIP versus checkpoint for the headline run (ntKBz):
    retrieval and zero-shot rise and plateau, linear probing climbs and
    levels off, and the reported checkpoint (dashed line, controller convergence)
    lies on the early plateau rather than at the peak.}
  \label{fig:curves}
\end{figure}

Finally, we analyse how performance evolves over training.
Metrics rise and then plateau, and the best checkpoints typically occur well
before the maximum number of epochs; we therefore select reported checkpoints by
the controller-convergence rule of \autoref{sec:dynamics}.
Per-checkpoint comparisons on linear probing are limited by the run-to-run
variance discussed in \autoref{sec:limitations} (Figure~\ref{fig:curves}).

\begin{table}[t]
  \centering
  \small
  \caption{%
    Training duration (gains over CLIP).
    Extending from 2 to 4 epochs does not improve performance ---
    zero-shot and linear probing decline, while retrieval is unchanged
    within run-to-run noise --- so we train for 2 epochs by default.}
  \label{tab:duration}
  \begin{tabular}{lrrr}
    \toprule
    \rowcolor{tableheader}
    Epochs & ZS & Retrieval & LP \\
    \midrule
    2 & $+2.00$ & $+1.54$ & $+7.30$ \\
    4 & $+0.03$ & $+1.59$ & $+3.78$ \\
    \bottomrule
  \end{tabular}
\end{table}

\section{Analysis}
\label{sec:analysis}

\subsection{Controller dynamics and learning-rate stability}
\label{sec:dynamics}

\paragraph{A fixed weight is inert at web scale.}
The KALE controller exists to correct a loss imbalance that emerges specifically
at CC12M scale.
Under the initial weight $\la^{(0)}=0.5$ (the value KUEA uses on ImageNet
\citep{KUEA}), the alignment term contributes only about $0.2\%$ of the clean
term during warmup ($\rho\approx2{\times}10^{-3}$), falling to
${\sim}10^{-5}$ at individual steps.
At this magnitude the alignment gradient is effectively negligible and
fine-tuning reduces to merely preserving the original CLIP features.
This imbalance is far milder on curated ImageNet-1K, which is why a fixed weight
suffices there but not on noisy web-scale data.

\paragraph{Rebalancing requires a large, non-transferable weight.}
To drive the weighted ratio $\rho$ toward its target $\tau\approx0.8$, the
controller increases $\la$ by roughly $7{,}000\times$ to $47{,}000\times$ over
training (from $0.5$ to ${\sim}3{\times}10^{3}$--$2{\times}10^{4}$), and the
required magnitude varies with learning rate and schedule.
No single fixed weight reproduces this behaviour: the balancing value is several
orders of magnitude larger than the ImageNet setting and is
configuration-dependent, which motivates an adaptive controller rather than a
hand-tuned constant.

\paragraph{The controller equilibrates rather than diverging.}
In stable runs, $\la$ rises steeply, peaks early (around step~2{,}400), and
then declines as the student aligns and the alignment loss shrinks.
The L2 drift from the original CLIP embeddings follows the same arc, peaking at
$\approx4$--$6$ before receding to $\approx3$, so the representation remains
bounded rather than being pushed monotonically away from CLIP.
This matches the fixed-point analysis of \autoref{sec:kale}: the controller
tracks $\la^\star\propto\lhat{clean}/\lhat{align}$, which peaks when the clean
loss (CLIP drift) peaks and then falls as drift recedes --- so the penalty weight
tracking it peaks and declines in step with the drift, rather than accumulating
without bound.

\paragraph{Learning rate sets the stability envelope.}
A high peak learning rate ($1{\times}10^{-4}$) is needed for the optimiser to
track the escalating alignment pressure, but the regime has a sharp upper
boundary: at $2{\times}10^{-4}$ training collapses, with drift running away
($\gtrsim15$), the embedding effective rank crashing (from $\approx106$ to
$\approx27$), and zero-shot accuracy falling far below the CLIP baseline.
The schedule matters in the opposite direction: a flat learning rate lets drift
grow unchecked throughout training (L2 rising from $3.6$ to $6.2$) and degrades
downstream performance, whereas cosine decay with a moderate floor reins drift
back in.
In this single-run sweep, a decaying schedule with a non-zero floor --- not a
flat one --- is associated with the most stable dual-loss optimisation
(Table~\ref{tab:floor}).
We do not claim a single optimal floor value: across configurations, differences
between moderate floors fall within the run-to-run variance reported in
\autoref{sec:limitations}.

\begin{figure*}[t]
  \centering
  \includegraphics[width=\linewidth]{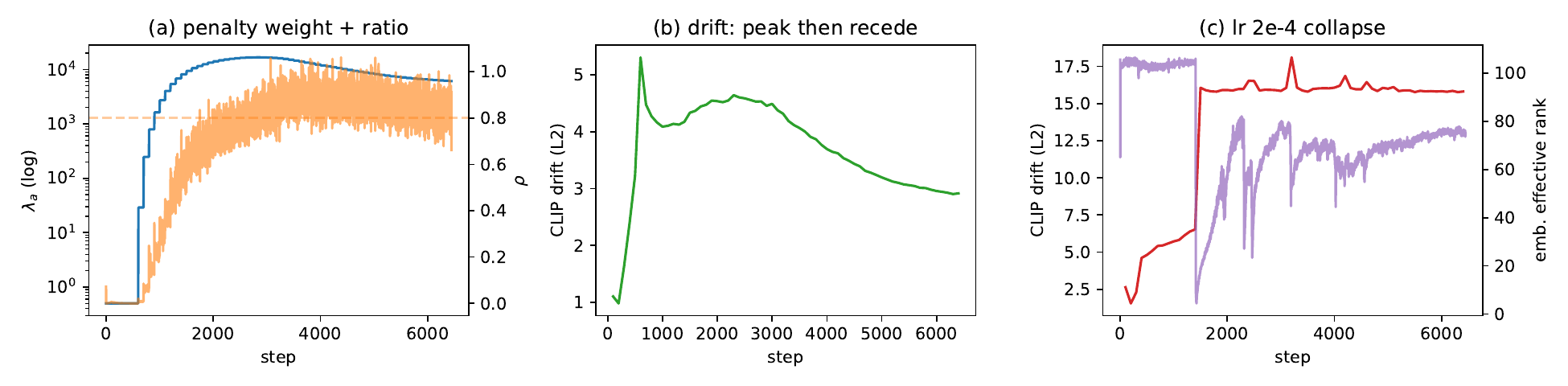}
  \caption{%
    Controller dynamics for the headline run.
    (a)~The penalty weight $\la$ (log scale) escalates by
    ${\sim}10^4{\times}$ and ratio $\rho$ converges to target $\tau{=}0.8$,
    then $\la$ peaks and declines.
    (b)~CLIP drift peaks early and recedes.
    (c)~At learning rate $2{\times}10^{-4}$ training collapses --- drift runs
    away and the embedding effective rank crashes.}
  \label{fig:dynamics}
\end{figure*}

\subsection{Domain breadth}
\label{sec:domain}

We had anticipated that the broader domain coverage of CC12M, relative to curated
ImageNet, might yield gains on structured or out-of-distribution tasks such as
remote sensing, texture recognition, and spatial reasoning.
The evaluation does not support this.
Relative to the ImageNet-trained KUEA baseline, web-scale training does not
produce broad-domain improvement: remote sensing (EuroSAT) regresses consistently
(\autoref{sec:results}), the CLEVR tasks are high-variance
(\autoref{sec:limitations}), and the per-dataset picture is mixed rather than
uniformly broader.
We therefore make no claim of improved domain breadth from web-scale training;
the contribution of KALE is in making web-scale alignment feasible and stable
(\autoref{sec:dynamics}), not in delivering wider downstream competence than a
curated-data baseline.

\subsection{PCAM}
\label{sec:pcam}

PCAM \citep{PCAM} is a balanced binary histopathology task (patch-level detection
of metastatic tissue) far from natural-image distributions, where CLIP sits near
the $50\%$ chance floor ($52.0\%$).
KALE improves it consistently --- every selected checkpoint beats CLIP --- though the gain is highly volatile, swinging more
than ten points between adjacent checkpoints and across runs, so we treat its
magnitude cautiously.
We hypothesise that any genuine component reflects DINOv2's texture-rich features
transferring to CLIP --- PCAM is a tissue-texture and morphology task --- rather
than medical content in the web-crawled CC12M data, which contains negligible
histopathology; this is consistent with KUEA's attribution of its own gains to
DINOv2's fine-grained perception.
A controlled test, correlating the per-checkpoint PCAM gain with alignment
strength, is left to future work.

\section{Limitations and Future Work}
\label{sec:limitations}

Our headline numbers come from our best single training run, and linear probing
shows substantial run-to-run variance.
Across seven identically-configured replicates --- which share an identical
learning-rate schedule through the reported checkpoint (verified) --- the
linear-probing average has a standard deviation of roughly two points,
concentrated almost entirely in the CLEVR tasks (Figure~\ref{fig:variance}),
while zero-shot and retrieval are stable (standard deviation below half a point).
We trace this to a training pipeline run without a fixed random seed or
deterministic kernels, so run-to-run nondeterminism is uncontrolled.
We therefore report the best run together with the run-to-run standard deviation
over these replicates, and base our conclusions on the metrics that are stable
across runs (retrieval, zero-shot, and SVHN linear probing); the CLEVR numbers
in particular should be read as high-variance.
Establishing CLEVR effects firmly would require multiple seeded runs per
configuration, which we leave to future work.

Our comparison to KUEA also isolates the method only partially.
We show directly that the alignment term is inert under a fixed weight at CC12M
scale and that the controller restores it (\autoref{sec:dynamics}), but the
fixed-weight baseline was not evaluated downstream at this scale, so we cannot
report a head-to-head downstream comparison between fixed-weight and KALE training
on identical web-scale data.
An eval-only experiment on fixed-weight CC12M checkpoints would close this gap
and let us attribute downstream differences to the controller rather than to the
change in training data relative to KUEA\@.

Our KALE controller is sensitive to the choice of learning rate and controller
hyperparameters: small deviations from the effective operating regime can lead
to meaningful performance regressions.
Future work could explore controllers parametrised in terms of relative training
progress or loss ratios rather than raw step counts, which may yield more robust
behaviour across different dataset sizes and batch configurations.

We observe a persistent trade-off between linear probing and retrieval performance
across checkpoints and configurations.
No single model simultaneously maximises both objectives, suggesting that the
alignment and clean losses encode partially competing pressures.
A natural extension would be to investigate multi-objective optimisation
formulations or Pareto-aware checkpoint selection strategies that explicitly
balance these downstream goals.

Our experiments focus on a 3.3M random subset of CC12M, which, while larger and
noisier than ImageNet-1K, is still a fraction of the data available for
web-scale training.
Scaling to the full CC12M or larger datasets such as DataComp \citep{DataComp}
or LAION \citep{LAION} would test whether the KALE controller remains effective
and whether performance continues to improve with data scale.

We have not explored alternative kernel functions, loss formulations, or teacher
models beyond the polynomial kernel and DINOv2 used in KUEA \citep{KUEA}.
Different kernels may offer better gradient properties at scale, and stronger or
domain-specialised teachers could further improve alignment quality.
We leave systematic exploration of these design axes to future work.

Finally, our evaluation is limited to zero-shot classification, linear probing,
and image--text retrieval.
Extending the analysis to dense prediction tasks, visual question answering, or
integration with multimodal LLMs would provide a more complete picture of how
KALE alignment affects downstream utility.

\section{Conclusion}
\label{sec:conclusion}

Kernel-based CLIP$\to$DINOv2 alignment works with a fixed loss weight on curated
ImageNet \citep{KUEA} but breaks down on noisy, web-scale data, where the
alignment term becomes too small to influence training.
A dynamic penalty-weight controller that adaptively rebalances the two losses
restores it, making alignment on a multi-million-sample CC12M subset stable.

Relative to the original CLIP model, the aligned encoder preserves image--text
retrieval, improves zero-shot classification, and reproducibly improves SVHN
linear probing; relative to the ImageNet-trained KUEA checkpoint it reaches a
different downstream operating point --- retrieval preserved, SVHN linear probing
improved, the standard zero-shot average improved over CLIP
(+2.00 vs.\ KUEA +1.29 \citep{KUEA}) --- achieved on noisy CC12M
without per-dataset weight tuning.
The contribution is making
kernel-based alignment trainable at this scale.
The accompanying training-dynamics analysis shows that the alignment term is
effectively inert under a fixed weight at this scale, and that stable
optimisation requires a bounded learning rate and a decaying schedule with a
moderate floor.
More broadly, the same control may help wherever a small-magnitude auxiliary
objective must stay influential against a dominant loss, beyond vision--language
alignment.

\bibliographystyle{icml2025}
\bibliography{refs}

\end{document}